\documentclass[letterpaper, 10 pt, conference]{ieeeconf}  
\usepackage{graphicx} 
\usepackage{amsmath}
\usepackage{amssymb}
\usepackage{booktabs}
\usepackage{comment} 
\usepackage{url}

\IEEEoverridecommandlockouts                              

\title{\LARGE \bf
Tree-of-Thoughts Reasoning for Text-to-Image In-Context Learning
}

\author{Stepanida Alekseeva$^{1}$, Jenifer Kalafatovich$^{1}$ and Seong-Whan Lee$^{1,*}$%
\thanks{This work was supported by the Institute of Information \& Communications Technology Planning \& Evaluation (IITP) grant funded by the Korea government (MSIT) (No. RS-2019-II190079, Artificial Intelligence Graduate School Program, Korea University), and by the Information Technology Research Center (ITRC) support program (No. IITP-2026-RS-2024-00436857).}%
\thanks{$^{1}$Department of Artificial Intelligence, Korea University, Seoul, Korea. e-mail:
{\tt\small alexeeva.stepanida@gmail.com}; {\tt\small jenifer@korea.ac.kr}; {\tt\small sw.lee@korea.ac.kr}}%
\thanks{$^{*}$Corresponding author.}%
}

\begin{document}

\maketitle

\thispagestyle{empty}
\pagestyle{empty}

\begin{abstract}

In text-to-image in-context learning (T2I-ICL), a model has to infer a latent compositional pattern from few-shot demonstrations for generating a query image. Recent studies show that state-of-the-art multimodal large language models struggle with this setting, particularly due to limited compositional reasoning and sensitivity to prompt construction.
In this work, we propose a Tree-of-Thoughts (ToT) reasoning framework for T2I-ICL that introduces a multi-stage reasoning and selection layer that generates, evaluates, and selects among multiple candidate hypotheses before constructing the final prompt for image synthesis. By exploring alternative reasoning branches and selecting a coherent interpretation, the proposed approach mitigates prompt ambiguity and compositional errors.
We implement the proposed approach in a complete ToT-T2I-ICL inference pipeline and evaluate it on the CoBSAT benchmark. Both qualitative and quantitative results show that structured multi-branch reasoning leads to more consistent and semantically aligned image generation compared to baseline and Chain-of-Thought prompting strategies, without any additional training or fine-tuning. \textit{Note:} The code is publicly available at \url{https://github.com/Pandastep/ToT-T2I-ICL}.

\end{abstract}


\section{Introduction}
Text-to-image (T2I) generation aims to synthesize images from natural language descriptions, enabling intuitive human-AI interaction and controllable visual content creation. Recent advances in T2I generation have enabled high-quality image synthesis from natural language descriptions \cite{rombach2022ldm, ramesh2021dal}. Despite strong visual fidelity, modern models still struggle with compositional generalization, particularly in correctly binding objects and attributes \cite{liu2022composable}.

 In-context learning (ICL) enables models to adapt to new tasks and patterns at inference time without parameter updates, making it a flexible and efficient alternative to task-specific training. While ICL has been widely adopted in multimodal large language models (MLLMs) for tasks like captioning and visual question answering \cite{alayrac2022flamingo, zhou2023contextdiffusion}, its application in T2I remains underexplored. Recent benchmarks such as CoBSAT reveal that MLLMs face  notable difficulties in T2I-ICL, indicating limited capacity for structured compositional reasoning \cite{zeng2024cobsat}.

In this paper, we address this limitation by introducing a Tree-of-Thoughts (ToT) reasoning framework for T2I-ICL. Instead of relying on a single reasoning trajectory as in previous methods, the proposed approach explores multiple candidate interpretations, evaluates them with respect to the in-context examples, and selects a coherent hypothesis for image generation. This structured reasoning process improves robustness to ambiguity and reduces compositional errors. Our method operates entirely at inference time and does not require model fine-tuning. Specifically, the model first analyzes the input demonstrations, then constructs a prompt through ToT reasoning, and finally generates an image based on the resulting prompt. This separation allows us to isolate the effect of reasoning on generation quality. 

The main contributions of this work are as follows:
(i) We propose a ToT reasoning framework for T2I-ICL that enables multi-branch hypothesis exploration and selection.
(ii) We design an inference-time pipeline that separates reasoning from image generation without additional training.
(iii) We demonstrate that multi-branch reasoning improves compositional generalization and image--text alignment on the CoBSAT benchmark compared to baseline and Chain-of-Thought (CoT) approaches.

\section{Related works}
\subsection{Text-to-Image Generation and Controllability}

Diffusion-based models, particularly Latent Diffusion Models (LDMs), dominate modern T2I generation by performing denoising in a compressed latent space \cite{rombach2022ldm}. Recent studies have explored diffusion interpretability by analyzing visual concepts and attention behavior during denoising \cite{park2024explaining}, while multi-concept personalization improves controllable T2I generation in complex scenes \cite{woo2025flipconcept}. Other approaches improve controllability through structured representations and additional conditioning signals, such as RANNI \cite{feng2024ranni} and VersaGen \cite{versagen2025}. However, these methods do not address compositional rule induction from multimodal demonstrations in T2I-ICL settings.

\subsection{In-Context Learning for Text-to-Image Generation}

ICL enables models to adapt to new tasks from a small set of demonstrations without parameter updates \cite{dong2024iclsurvey}. Related few-shot learning studies show that geometric constraints, self-augmentation, and spatial reasoning improve generalization from limited examples \cite{jung2020fewshot,seo2021selfaugmentation,kim2021spatial}. Context-based activity recognition also highlights the role of contextual structure in visual understanding \cite{kim2018context}.

Zeng et al. introduced the CoBSAT benchmark and demonstrated that current MLLMs struggle with T2I-ICL \cite{zeng2024cobsat}. ImageGen-CoT further improves performance by generating explicit reasoning chains before image synthesis \cite{liao2025imagegen}. However, existing approaches rely on linear reasoning and do not explicitly explore multiple candidate hypotheses.

\subsection{Visual In-Context Learning with Diffusion Models}

Several works aim to enable in-context behavior directly within diffusion models. Prompt Diffusion performs task transfer using example image pairs and textual guidance \cite{wang2023icl}, while Analogist enables visual ICL through attention-based mechanisms and semantic guidance \cite{wu2024analogist}. Related vision studies on image retrieval, face recognition, action recognition, and video text extraction show the importance of robust visual representation under varied visual conditions \cite{lee2012pillid,kang2014nighttime,roh2010view,lim2000text}. These methods primarily focus on recognition or image editing rather than compositional concept induction from text-image demonstrations.

\subsection{Reasoning for Image Generation: CoT and ToT}

CoT \cite{wei2022cot} and ToT \cite{yao2023tot} improve multi-step reasoning by generating intermediate reasoning steps and exploring multiple candidate solutions. However, ToT-style reasoning has not been systematically explored for T2I-ICL, where the model must infer and transfer compositional patterns from multimodal demonstrations.

\section{Methodology}
\subsection{Problem Setup} We study text-to-image in-context learning (T2I-ICL), where a model receives a small set of image-text demonstration pairs and a textual query, and must generate an image that correctly applies the demonstrated compositional pattern to the query. 

Formally, an ICL task is defined by: (i) a set of $N$ demonstration examples
\begin{equation}
\mathcal{E} = \{(t_i, i_i)\}_{i=1}^{N},
\end{equation}
where $t_i$ is a textual description and $i_i$ is the corresponding image, and $N$ denotes the number of in-context demonstration examples, and (ii) a query text $t_q$ specifying a target concept.

The challenge lies in compositional generalization; the target concept in the query may not appear in the demonstrations and must be correctly integrated into the learned visual pattern.


\begin{figure*}[t]
  \centering
  \includegraphics[width=1\textwidth]{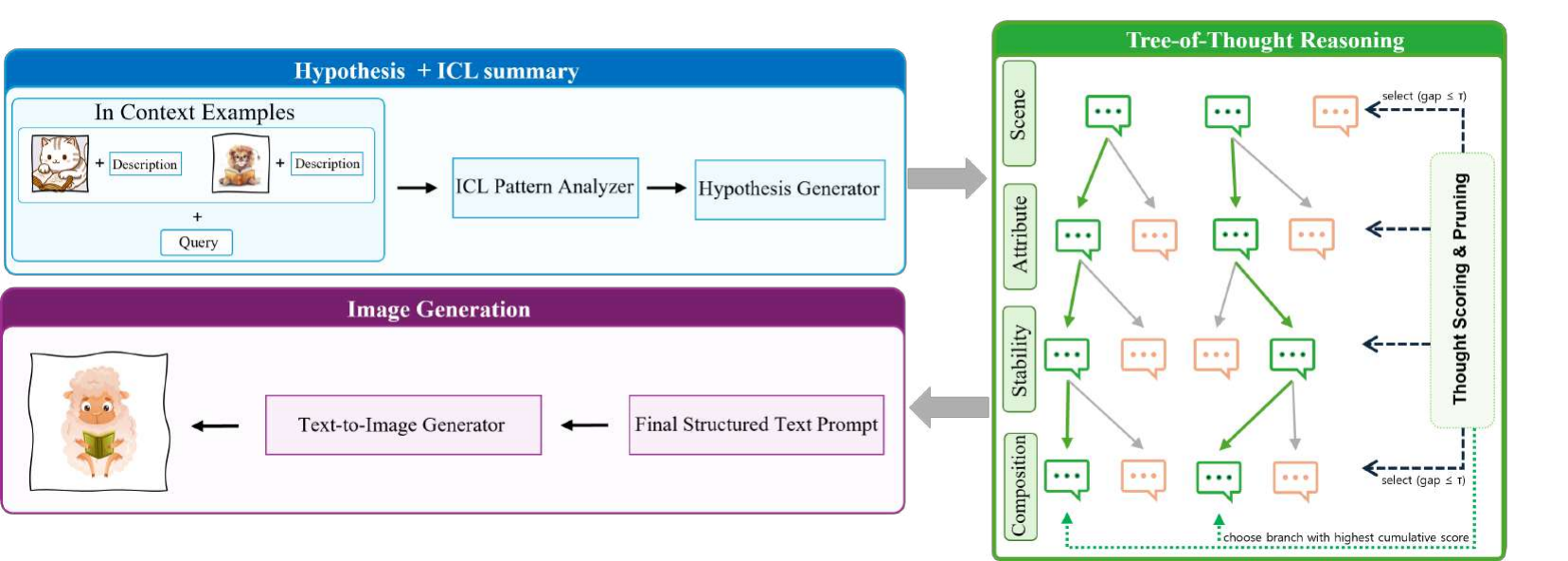}
  \vspace{-15pt}
    \caption{Overview of the proposed ToT-T2I-ICL pipeline.
    The method performs stage-wise Tree-of-Thoughts reasoning over candidate interpretations derived from in-context examples and a query.
    Multiple reasoning branches are explored and pruned at each stage, and a single coherent path is selected to construct the final prompt for text-to-image generation. Green nodes indicate candidate thoughts retained after scoring and adaptive branch selection based on score differences.}
    \label{fig:tot_pipeline}
\end{figure*}

\subsection{Pipeline Overview} We propose ToT-T2I-ICL, a Tree-of-Thoughts reasoning pipeline for text-to-image in-context learning. Instead of collapsing all reasoning into a single textual prompt, our approach explicitly decomposes prompt construction into a structured, multi-stage reasoning process that explores multiple alternative interpretations before selecting a final prompt for image generation.

As illustrated in Fig.~\ref{fig:tot_pipeline}, the pipeline operates in three clearly separated phases: an input representation and pattern analysis phase, which extracts task-relevant structural information from the in-context demonstrations and the query; a reasoning phase, where candidate interpretations are generated and evaluated using a multimodal language model, and an image generation phase, where a fixed text-to-image generator synthesizes the final image based solely on the selected prompt.





\subsubsection{Phase I - Input Representation and Pattern Analysis} 

The model first analyzes the input demonstrations to infer task-relevant structure directly from data. Specifically, it extracts:
(i) invariant elements shared across demonstrations,
(ii) varying components that change between examples,
and (iii) the query entity.

\noindent Based on this analysis, a structured hypothesis is constructed. The hypothesis includes: a natural-language transfer rule describing how the pattern should be applied,
and a set of constraints that restrict invalid transformations. These constraints enforce consistency with the demonstrations by preventing unsupported modifications, such as introducing unrelated objects or violating the inferred transformation. This design removes reliance on dataset-specific metadata and enables generalization across tasks.


\subsubsection{Phase II - Tree-of-Thoughts Reasoning} The core of our approach is a fixed-depth ToT reasoning process that explores alternative interpretations of the target image before selecting a final prompt.

\noindent \textbf{Stage Definition. } 
Reasoning is performed as a structured search process over multiple candidate thoughts. The generation is decomposed into four semantically distinct stages:
\textit{Scene} (high-level interpretation),
\textit{Attribute} (visible transformation),
\textit{Stability} (coherence and recognizability),
and \textit{Composition} (presentation and layout).

\noindent At each stage, the model generates multiple candidate thoughts for each active reasoning branch (3 candidates for early stages and 2 for later stages). Each branch maintains a sequence of previous thoughts, which is used both to guide subsequent generation and to ensure consistency during evaluation. This enables incremental refinement rather than independent step-wise reasoning.

\noindent The four-stage decomposition is a design choice to control ambiguity in T2I generation. Early stages (Scene and Attribute) establish semantic grounding, while later stages (Stability and Composition) refine consistency and layout, enabling a coarse-to-fine process. Scene and Attribute follow a global-to-local order and are not fully interchangeable, as reversing them can increase ambiguity due to missing context. Stability and Composition are orthogonal and could be interchanged; we fix their order for consistency and reduced prompt instability during multi-branch reasoning.


\noindent \textbf{Branching Mechanism.}
Reasoning is formulated as a multi-branch search process over stage-wise thoughts. At each stage, a set of active branches is maintained, and the model generates candidate thoughts for each branch.

\noindent Each branch is represented as:
\begin{equation}
b = \langle \text{thoughts}, \text{score\_sum} \rangle,
\end{equation}
where \texttt{thoughts} is the sequence of selected stage-specific thoughts and \texttt{score\_sum} is the accumulated score.

\noindent Each candidate extends the current branch with one additional thought and is subsequently evaluated and filtered by the scoring and pruning procedure.


\noindent \textbf{Scoring, Branch Selection, and Pruning.}
At each stage, the MLLM generates multiple candidate thoughts for every active branch. Each candidate thought $  t  $ receives a scalar score
\begin{equation}
s(t) = \sum_{k} w_k f_k(t) - p_{\text{leap}}(t) - p_{\text{red}}(t),
\label{eq:scoring}
\end{equation}
where $f_k(t) \in [0,1]$ denotes normalized heuristic components and $w_k=\{0.24,0.22,0.18,0.14,0.12,0.10\}$ denotes the corresponding weights for the six criteria defined below. The weights are fixed and manually set based on preliminary experiments to balance semantic relevance and structural consistency, ensuring reproducibility.
The scoring function evaluates six complementary criteria:
\begin{itemize}
\item \textit{Query anchoring}: how directly the thought addresses the query concept;
\item \textit{Entity preservation}: consistency with the invariant elements from the hypothesis;
\item \textit{Stage fidelity}: semantic appropriateness to the current stage (scene, attribute, stability, or composition);
\item \textit{Rule consistency}: alignment with the inferred transformation rule;
\item \textit{Constraint consistency}: compliance with explicit hypothesis constraints;
\item \textit{Linguistic quality}: conciseness, specificity, and absence of vague language.
\end{itemize}
Two deterministic penalties are subtracted to explicitly handle common failure modes in compositional reasoning:
\begin{itemize}
\item \textit{Unsupported leap penalty} ($  p_{\text{leap}} \in [0,0.5] $): for abrupt or weakly justified semantic shifts away from the demonstrated pattern;
\item \textit{Redundancy penalty} ($  p_{\text{red}} \in [0,0.5] $): for high semantic overlap with prior thoughts in the same branch.
\end{itemize}
All components rely solely on lexical matching, keyword detection, and lightweight structural heuristics. We do not use any learned evaluators or external models. Lexical matching refers to simple token-level overlap and keyword-based comparison between the thought, query, and hypothesis elements.
Candidates are ranked by score. Pruning follows a simple beam-search rule with threshold $  \tau  $ : if the score gap between the top two candidates is $  \leq \tau  $, both are retained (up to beam width 2); otherwise, only the highest-scoring candidate proceeds. This mechanism preserves plausible ambiguity when candidates are close while aggressively pruning weak branches. Surviving thoughts extend their parent branch, and branch scores are accumulated as the sum of selected thought scores.
After the final stage, the branch with the highest cumulative score is selected as the winning reasoning path.
\subsubsection{Phase III - Image Generation} 
 \noindent \textbf{Final Prompt Construction.} The final prompt is constructed from the selected reasoning path by combining the target subject, inferred invariants, and retained stage-specific visual phrases. A rule-based filter removes meta-language, demonstration references, overly generic or reference-only phrases, and redundant fragments, followed by normalization and deduplication to obtain a text-to-image-compatible prompt.

\noindent The resulting prompt is concise, visually grounded, and aligned with the inferred transformation pattern. This prompt is then used by a frozen text-to-image generator to produce the final image.


\noindent \textbf{Image Generation.} The finalized structured prompt obtained after ToT reasoning and branch selection process is passed to a T2I diffusion model to generate the output image.


\section{Results}


\begin{table}[t]
\centering
\caption{Overall comparison using CLIP similarity and Constraint Satisfaction Rate (CSR) across all samples. Higher values are shown in bold.}
\label{tab:overall_results}
\begin{tabular}{lcc}
\hline
Method & CLIP & CSR \\
\hline
Baseline & 0.287 $\pm$ 0.032 & 0.508 $\pm$ 0.336 \\
CoT      & 0.302 $\pm$ 0.031 & 0.547 $\pm$ 0.341 \\
ToT      & \textbf{0.318 $\pm$ 0.030} & \textbf{0.775 $\pm$ 0.252} \\
\hline
\end{tabular}
\end{table}

\begin{table}[t]
\centering
\caption{Human evaluation results (21 participants, 20 samples). Values denote preference proportions.}
\label{tab:human_eval}
\begin{tabular}{lccc}
\hline
\textbf{Criterion} & \textbf{Baseline} & \textbf{CoT} & \textbf{ToT} \\
\hline
Example & 0.217 & 0.188 & \textbf{0.595} \\
Query   & 0.121 & 0.198 & \textbf{0.681} \\
Joint   & 0.157 & 0.188 & \textbf{0.655} \\
\hline
\end{tabular}
\end{table}

\setlength{\textfloatsep}{2pt}
\setlength{\floatsep}{2pt}
\setlength{\abovecaptionskip}{1pt}
\setlength{\belowcaptionskip}{5pt}
\begin{figure*}[t]
  \centering
  \includegraphics[width=1\textwidth]{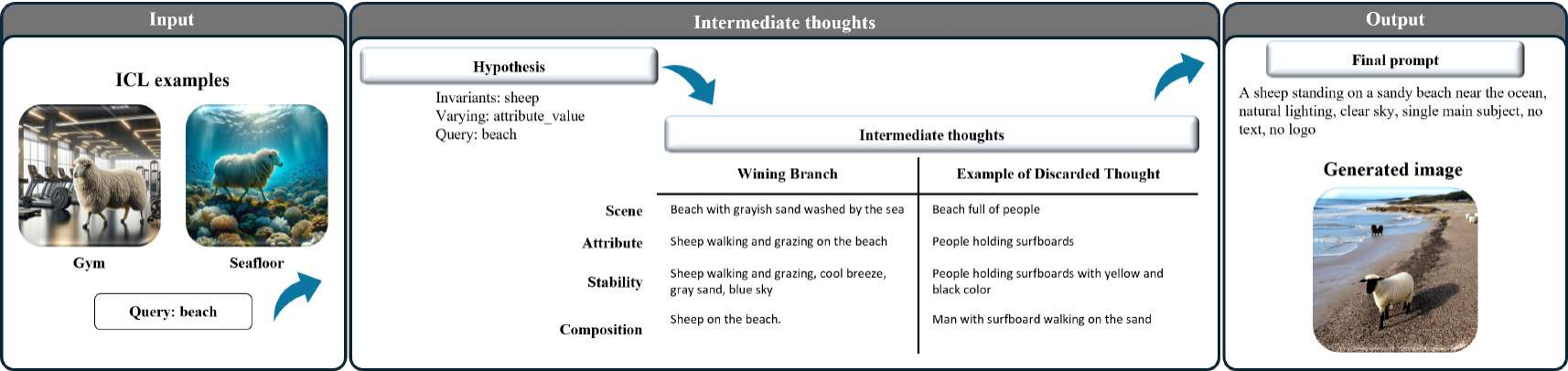}
  \caption{
Qualitative results for one example of ToT-based T2I-ICL. The model correctly transfers the concept from the provided ICL examples (sheep) to the given query (beach), producing a semantically consistent scene. Minor background artifacts do not affect interpretation.
}
    \label{fig:ex3}
\end{figure*}

\begin{figure*}[t]
  \centering
  \includegraphics[width=1\textwidth,height=0.27\textheight]{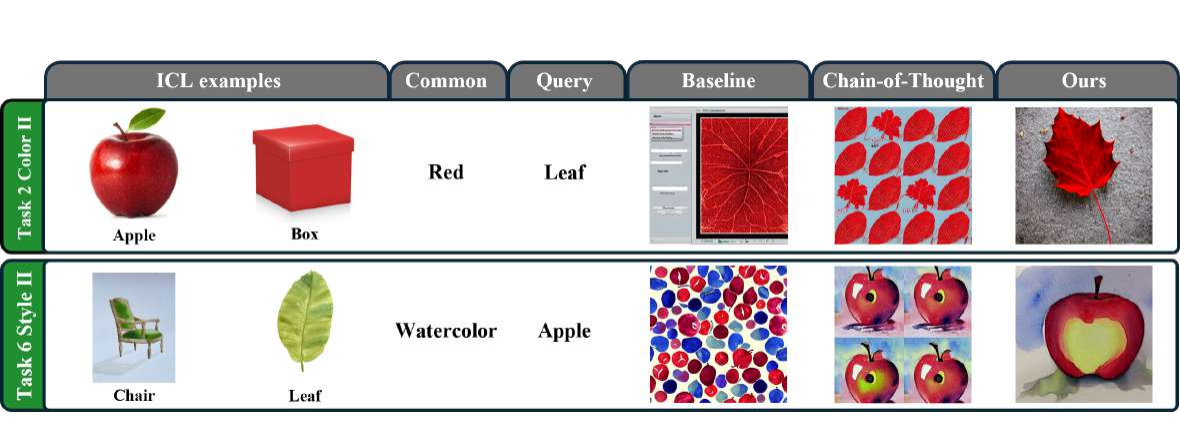}
  \vspace{-20pt}
    \caption{Comparison results of text-to-image in-context learning across two CoBSAT tasks. In the Color task (top), the baseline produces a noisy texture without a clear object, while CoT generates repeated patterns of the source examples rather than adapting to the query. In contrast, ToT correctly binds the target attribute and object, producing a coherent red leaf. In the Style task (bottom), the baseline again fails to form a meaningful object, and CoT introduces multiple inconsistent instances. ToT instead generates a single, well-structured apple with the desired watercolor style, demonstrating improved compositional reasoning and alignment.}
    \label{fig:exp1}
\end{figure*}

\subsection{Experiments} 
We evaluate the proposed ToT reasoning framework on the CoBSAT benchmark, which consists of ten compositional T2I-ICL tasks. Each task requires the model to infer a target visual concept from a small set of image--text demonstration examples and a textual query.

Our system follows a two-stage inference pipeline that explicitly separates reasoning from image synthesis. The SEED-LLaMA \cite{ge2023making} multimodal model is used as the reasoning backbone to perform in-context reasoning over the demonstration examples and the query, producing a structured final text prompt. The generated prompt is then passed to Stable Diffusion \cite{rombach2022ldm}, which generates the corresponding output image. This separation allows us to isolate the effect of reasoning strategies on generation quality. 

We compare three reasoning strategies: Baseline, which constructs the final prompt directly without explicit reasoning;
Chain-of-Thought, which generates a single linear reasoning trace prior to prompt construction; and 
 Tree-of-Thoughts, which explores multiple alternative reasoning paths across structured stages and selects a final reasoning branch for prompt construction. All methods use identical in-context examples, identical image generation parameters, and fixed random seeds. No fine-tuning or parameter modification is applied to either model. Therefore, performance differences reflect only the reasoning and prompt construction strategy. For each query, a single image is generated.

\subsection{Implementation Details} Our implementation uses a fixed four-stage reasoning depth, limited branching width per stage, score-based pruning with threshold $\tau = 0.08$, and a deterministic, rule-based hypothesis generation.
 
This configuration balances reasoning diversity and computational efficiency. In preliminary experiments, smaller values of $\tau$ led to unstable reasoning paths dominated by noisy or underspecified hypotheses, while larger values resulted in overly aggressive pruning and reduced diversity.

We therefore fix this value across all experiments without task-specific tuning, enabling stable and scalable evaluation on the full CoBSAT benchmark.

\subsection{Evaluation Metrics}

We evaluate the generated images using two complementary automatic metrics. We report the CLIP Score \cite{radford2021learning}, computed as the cosine similarity between the CLIP (ViT-B/32) image embedding and the query text embedding. While it provides a useful proxy for semantic alignment, it has limited sensitivity to fine-grained compositional correctness.

To more precisely measure structural fidelity in T2I-ICL, we introduce the Constraint Satisfaction Rate (CSR). For each sample, we define two task-specific constraints derived from the ground-truth CoBSAT task definition: (i) preservation of the invariant element inferred from the ICL examples, such as object or subject identity, and (ii) realization of the query-specific change. Depending on the task, the query-specific constraint corresponds to color, background, style, action, or texture transformation.

CSR is computed using CLIP-based ranking. For each constraint, the generated image is compared against one correct textual description and several incorrect variants. A constraint is considered satisfied if the correct description obtains the highest CLIP similarity score.

Formally, for a generated sample $i$ with constraint set $\mathcal{C}_i$,
\begin{equation}
\text{CSR}_i = \frac{1}{|\mathcal{C}_i|} \sum_{c \in \mathcal{C}_i} \mathbb{1}\Bigl[ \text{constraint } c \text{ is satisfied} \Bigr],
\end{equation}
where $\mathbb{1}[\cdot]$ is the indicator function. The final CSR is averaged across all generated samples.

All results are reported on 300 samples (10 tasks $\times$ 30 samples per task), with mean and standard deviation for both CLIP Score and CSR.

\subsection{Cross-Method Comparison and Task-wise Performance Analysis} Table~\ref{tab:overall_results} summarizes the overall CLIP and CSR results across all samples. Specifically, CLIP similarity is computed for each generated image–text pair within a task and then aggregated over all samples and tasks to obtain the reported statistics.

ToT achieves the best performance across both metrics, obtaining a CLIP similarity of 0.318 ± 0.030 and a CSR of 0.775 ± 0.252, outperforming both Baseline (0.287 / 0.508) and CoT (0.302 / 0.547). This indicates that ToT not only improves semantic alignment between generated images and queries but also more reliably satisfies structural constraints derived from the ICL examples.

Fig.~\ref{fig:ex3} presents a step-by-step example of the ToT pipeline. Given ICL demonstrations (sheep in a gym and underwater) and the query “beach”, the model correctly identifies “sheep” as the invariant element and the scene as the varying component. During the intermediate reasoning stage, the selected branch explicitly constructs a coherent scene (beach), assigns consistent attributes (visible, grazing sheep), and enforces stability and composition constraints (single main subject). The resulting prompt leads to a semantically correct image of a sheep on a beach, demonstrating successful concept transfer from the ICL examples while preserving object identity.

Fig.~\ref{fig:exp1} provides a qualitative comparison that illustrates the generation results obtained from identical in-context examples and queries, where the difference lies in the reasoning strategy used to construct the final prompt. The Baseline method often produces scattered or inconsistent outputs, while CoT improves overall coherence but still introduces artifacts or compositional inconsistencies.
In contrast, the proposed ToT method yields stronger attribute–object binding and better preservation of the demonstrated patterns.

Table~\ref{tab:human_eval} reports the human evaluation results with 21 participants over 20 samples selected from the evaluated CoBSAT tasks, covering color, background, style, action, and texture transformations. For each sample, participants viewed the ICL examples, the query, and anonymized Baseline/CoT/ToT outputs in randomized order. They selected the best result according to consistency with ICL examples (Example), alignment with the query (Query), and overall quality (Joint). ToT is preferred in the majority of cases across all criteria, achieving $59.5\%$, $68.1\%$, and $65.5\%$ preference rates, respectively.  A one-sided binomial test against the three-way chance level of $1/3$ shows that the ToT preference is significant for all criteria ($p<0.001$). This suggests that the improvements observed in CLIP similarity and CSR also translate into perceptual quality gains.

\setlength{\textfloatsep}{9 pt}
\begin{table}[t]
\centering
\caption{Ablation study on the effect of Tree-of-Thoughts branch width $B$.
Reasoning score (s), CLIP similarity, and CSR are reported as mean $\pm$ std.}
\label{tab:ablation_branch}

\begin{tabular}{c|ccc}
\hline
\textbf{$B$} & \textbf{Reasoning Score (s)} & \textbf{CLIP} & \textbf{CSR} \\
\hline
1 & 0.489 $\pm$ 0.145 & 0.3206 $\pm$ 0.0314 & 0.8833 $\pm$ 0.2151 \\
2 & 0.649 $\pm$ 0.109 & \textbf{0.3262 $\pm$ 0.0348} & 0.8833 $\pm$ 0.2151 \\
3 & 0.700 $\pm$ 0.100 & 0.3064 $\pm$ 0.0442 & \textbf{0.9000 $\pm$ 0.2034} \\
5 & \textbf{0.723 $\pm$ 0.061} & 0.3060 $\pm$ 0.0334 & 0.8833 $\pm$ 0.2151 \\
\hline
\end{tabular}
\end{table}

\setlength{\textfloatsep}{9 pt}
\begin{table}[t]
\centering
\caption{Sensitivity analysis of the scoring weights in Eq.~\eqref{eq:scoring}.
CLIP similarity and CSR are reported as mean $\pm$ std.}
\label{tab:weight_sensitivity}

\begin{tabular}{l|cc}
\hline
\textbf{Weight Setting} & \textbf{CLIP} & \textbf{CSR} \\
\hline
Equal & 0.3027 $\pm$ 0.0411 & 0.9167 $\pm$ 0.1895 \\
w/o Query Anchoring & \textbf{0.3043 $\pm$ 0.0394} & \textbf{0.9333 $\pm$ 0.1729} \\
w/o Constraint Consistency & 0.3021 $\pm$ 0.0400 & 0.9167 $\pm$ 0.1895 \\
Original & 0.3032 $\pm$ 0.0404 & \textbf{0.9333 $\pm$ 0.1729} \\
\hline
\end{tabular}
\end{table}

\subsection{Ablation and Sensitivity Analysis}

To analyze the contribution of the proposed Tree-of-Thoughts (ToT) framework, we conduct ablation studies on a representative CoBSAT subset covering \textbf{Color I} (object color inference with fixed object identity), \textbf{Background I} (background modification while preserving foreground objects), \textbf{Style I} (style transformation while preserving object identity), \textbf{Action I} (action-centric inference involving animal subjects), and \textbf{Texture I} (texture transformation while preserving object identity). All configurations use the same multimodal backbone, in-context examples, generation hyperparameters, scoring function, selection threshold $\tau$, fixed samples, and random seed.

Table~\ref{tab:ablation_branch} analyzes the effect of candidate generation width $B \in \{1,2,3,5\}$, which controls the number of candidate reasoning hypotheses generated at each stage. $B=1$ corresponds to single-chain reasoning, while larger values enable exploration of multiple alternative reasoning paths before selection. Increasing $B$ consistently improves the internal reasoning score, with the largest gain from $B=1$ to $B=2$ and smaller gains thereafter. However, image-level metrics do not improve monotonically: CLIP is highest at $B=2$, while CSR is highest at $B=3$. This indicates that better intermediate reasoning does not always directly translate into better final image generation, likely due to branch-selection noise and the limited controllability of the frozen image generator. We therefore use $B=3$ as a balanced setting between reasoning quality, constraint satisfaction, and computational cost.

To examine the sensitivity of the manually assigned weights in Eq.~(3), Table~\ref{tab:weight_sensitivity} compares the original weights, equal weights, removal of query anchoring, and removal of constraint consistency; the remaining weights are renormalized. CLIP similarity varies only within a narrow range of 0.3021--0.3043, while CSR remains between 0.9167 and 0.9333, indicating that the scoring function is robust to moderate weight changes and is not dominated by a single manually selected criterion.

\section{Conclusion and Future work}
We presented a Tree-of-Thoughts reasoning framework for text-to-image in-context learning. By decomposing prompt construction into interpretable stages and explicitly exploring multiple reasoning branches, our method enables multimodal models to better capture and transfer compositional patterns from a small set of demonstrations.

Experimental results on the CoBSAT benchmark show that the proposed ToT-T2I-ICL framework consistently outperforms both direct prompting and linear Chain-of-Thought strategies. Ablation studies confirm that branching improves internal reasoning quality, while gains in CLIP similarity are limited and non-monotonic, highlighting the value of multi-hypothesis exploration.
However, improvements in internal reasoning scores do not always translate into proportional gains in CLIP similarity, suggesting that standard image–text metrics may not fully capture fine-grained compositional correctness.

The main limitations of the current framework are the increased inference-time computational cost and evaluation on a controlled benchmark with limited task diversity.
Future work will focus on adaptive branching strategies that dynamically adjust the number of hypotheses, more robust scoring mechanisms, and scaling the approach to larger and more diverse open-domain T2I-ICL scenarios.

{\small
\bibliographystyle{IEEEtran}
\bibliography{egbib}
}
\end{document}